# A Practical Second-order Latent Factor Model via Distributed Particle Swarm Optimization

Jialiang Wang, Yurong Zhong, Weiling Li

**Abstract**—Latent Factor (LF) models are effective in representing high-dimension and sparse (HiDS) data via low-rank matrices approximation. Hessian-free (HF) optimization is an efficient method to utilizing second-order information of an LF model's objective function and it has been utilized to optimize second-order LF (SLF) model. However, the low-rank representation ability of a SLF model heavily relies on its multiple hyperparameters. Determining these hyperparameters is time-consuming and it largely reduces the practicability of an SLF model. To address this issue, a practical SLF (PSLF) model is proposed in this work. It realizes hyperparameter self-adaptation with a distributed particle swarm optimizer (DPSO), which is gradient-free and parallelized. Experiments on real HiDS data sets indicate that PSLF model has a competitive advantage over state-of-the-art models in data representation ability.

## I. Introduction

With the vigorous development of the Internet, especially the mobile Internet, the big bang data explosion is changing the world dramatically [1], e.g., e-shopping [2], e-library [2], short or long video media applications [3, 4], and news applications [5]. This phenomenon is accompanied by the creation of a by-product called data overload [6, 7]. To free people from this dilemma, Recommender System (RS), a tool that can filter out the information that is useful to people from massive amounts of data [8].

Normally, interactions between different users and different items can be represented by user-item rating matrices. However, the representation of these rating data in the matrices are extremely sparse. In other words, there are lots of missing data in the rating matrices due to it is impossible for one person to rate all items, and the same goes for an item. Although the rating matrices are ultra-high-dimensional and sparse (HiDS), the RS can still extract rich knowledge from them [9-11].

The family of Latent Factor (LF) models, the important parts of RS, show their power to represent interaction pattern by projecting high-level features into two low-rank matrices [41, 43-48]. Many researchers have a lot of research on LF models. For example, Luo *et al.* [12] proposed a nonnegative LF model based single-element update. Zhang *et al.* [13] proposed a novel distributed quantized ADMM to speed up and optimize the LF model. Liu *et al.* [14] proposed an efficient deep matrix factorization with review feature learning for RS. Wu et al. [15] proposed a robust LF model for precise representation of HiDS data.

Surprisingly, LF models have made good progress in recent years, however, these models mainly adopt first-order type optimizer, especially the family of vanilla stochastic gradient descent [16-21]. Although the first-order class algorithm can achieve good results on the LF models, there is no recognized first-order class optimization algorithm that can achieve good results in all extremely sparse data scenarios and choosing a suitable first-order optimizer is itself a hyperparameter. The reason is that such algorithms do not consider the local curvature at a certain point [22]. Hence, second-order LF model (SLF) is able to obtain better representation performance due to it assimilates second-order information, local curvature landscape, during the training process of LF model rather than first-order information only [23, 24]. However, the representation ability of the SLF model largely depends on the fine-tuning of the hyperparameters. Since we don't know the specific function of any model w.r.t hyperparameters. Therefore, gradient-free optimization techniques, like grid search, play an important role in this process [25]. Therefore, how to adaptively optimize the optimal or sub-optimal hyperparameters is SLF model is a critical issue worthy research.

According to prior research [25-27, 33], evolutionary algorithms or swarm intelligence algorithms are widely utilized to address this issue. The Particle Swarm Optimization (PSO), a kind of swarm intelligence algorithm, has been verified to be much research to handle gradient-free optimization problems including hyperparameter optimization. The serial PSO (SPSO) scheme is prone to fall into the predicament of low computational efficiency, so many researchers have made this progress, resulting in many variants of the standard PSO algorithm. However, because of high computation overhead, it does not work well exactly. Hence, optimizing the hyperparameters remains a substantial obstacle in designing SLF model in practice.

This paper proposes a PSLF model. This model can not only realize the self-adaption of SLF model's hyperparameters, thereby improving the representation performance of the SLF model, but also reduce the optimization time due to the use of a distributed parallel strategy. The main contributions of this paper include:

a) A PSLF model who can automatically determine its hyperparameters via DPSO optimizer;

b) Algorithm design and analysis of proposed; and

c) Empirical studies on three HiDS matrices generated by industrial applications.

The rest of this paper is organized as follows. Section II gives the preliminaries. Section III presents an PSLF model. Section IV provides the experimental results and analysis. Finally, Section V draws the conclusions.

## II. PRELIMINARIES

### A. Problem Statement

Given an arbitrary RS ratings dataset, a user-item rating score matrix $\mathbf{S} \in \mathbb{R}^{|U| \times |I|}$ can be formed, where $U$ denotes the user set, $I$ denotes the item set. Each element $s_{u,i}$ of $\mathbf{S}$ is the rating score of item $i \in I$ by user $u \in U$. $\mathbf{S}$ is a HiDS matrix subject to $|K| \ll |M|$ where $K$ and $M$ refer to the known and missing data of $\mathbf{S}$, respectively.

Given $K$ and $\mathbf{S}$, the purpose of LF model is to predict the missing data, the unknown rating score of $\mathbf{S}$. An LF model builds two low-rank LF matrices $\mathbf{P} \in \mathbb{R}^{|U| \times D}$ and $\mathbf{Q} \in \mathbb{R}^{|I| \times D}$, and realizes the approximation of $S$ by low-rank matrix factorization, $\hat{\mathbf{S}} = \mathbf{P}\mathbf{Q}^T$, where $D \ll \max\{|U|, |I|\}$.

Note that D denotes the size of LF dimension. It is based on the assumption that high-dimension features of users or items can be projected into a low-rank space for representation. In order to obtain $\mathbf{P}$ and $\mathbf{Q}$, an empirical loss function is required to measure the gap between the real rating score in K and the corresponding predicted value in $\hat{\mathbf{S}}$. To simplify empirical loss function, following [23, 24] map $\mathbf{P}$ and $\mathbf{Q}$ into a hyper-matrix, $\mathbf{X} \in \mathbb{R}^{(|U|+|I|) \times D}$. Then, an empirical loss function is as follows:

$$E(\mathbf{X}) = \frac{1}{2} \sum_{s_{u,i} \in K} \left( s_{u,i} - \sum_{d=1}^{D} x_{u,d} x_{(|U|+i),d} \right)^2 \quad (1)$$
$$s.t. \ \forall u \in U, i \in I, d \in \{1,...,D\},$$

where $s_{u,i}$, $x_{u,d}$ and $x_{i,d}$ denote the element in $\mathbf{S}$, $\mathbf{X_P}$, and $\mathbf{X_Q}$ respectively.

## III. PSLF MODEL

### A. A Second-Order LF Model

a. Base Components

The Hessian-free optimization [28], first proposed by Martens, a second-order method that no need to compute the Hessian $H_E(\mathbf{X})$ and the Hessian inverse $H_E^{-1}(\mathbf{X})$ w.r.t $\mathbf{X}$. The Hessian-free optimization can be mainly divided into two steps. First, it uses the conjugate gradient method (CG), an iterative method and can be regarded as between Newton-type method and steepest descent, to solve the linear system (3) without calculating the Hessian inverse. Theoretically, applying CG to solve (3) requires searching the orthogonal directions N times to obtain a Newton step size. But in Hessian-free optimization, instead of strictly applying CG, an inexact CG is used, reducing the computational overhead to $\Theta((|N| \times d)^2)$.

Second, in CG, $H_E(\mathbf{X})$ always coms in the form of Hessian-vector product, i.e., $H_E(\mathbf{X})\mathbf{v}$. In [28, 29], we can compute $\omega_E = H_E(\mathbf{X})\mathbf{v}$ without computing the full Hessian, where $\omega_E$ denotes $E$'s directional derivative along vector $v$. So, $\omega_E$ can be rewritten as follows:

$$\omega_E = H_E(\mathbf{X})\mathbf{v} = \frac{\partial}{\partial \varepsilon} \nabla E(\mathbf{X} + \varepsilon \mathbf{v}) \bigg|_{\varepsilon \to 0}. \quad (2)$$

In [29], (2) can be solved by R-operator. The Hessian-vector product, $\omega_E$, can be arrived as follows:

$$\omega_E = R\{\nabla E(\mathbf{X})\} = \left(R\left\{\frac{\partial}{\partial x_1}E(\mathbf{X})\right\},...,R\left\{\frac{\partial}{\partial x_{(|U|+|I|)\times D}}E(\mathbf{X})\right\}\right)^{\top}. \tag{3}$$

b. The Gauss-Newton Matrix Approximation

Since (1) is a non-convex and bilinear function, therefore, $H_E(\mathbf{X})$ may not be non-positive semi-definite. In other words, $H_E(\mathbf{X})$ may have negative eigenvalues, which does not meet the use conditions of CG. The Gauss-Newton matrix, an approximation of Hessian, is guaranteed to be positive semi-definite [23, 24, 28]. In practice, the Gauss-Newton matrix works better than the Hessian or other approximations of Hessian. Hence, $H_E(\mathbf{X})$ will be substituted by Gauss-Newton matrix, $G_E(\mathbf{X})$. Then (2) can be reformulated as follows:

$$G_E(\mathbf{X})\Delta\mathbf{X} + \nabla E(\mathbf{X}) = 0. \tag{4}$$

Before solving the linear system (4) with CG, to simplify the solution process, we only consider the following empirical loss function E(**X**) first.

$$E(\mathbf{X}) = \frac{1}{2}\sum_{s_{u,i}\in K}\left(s_{u,i} - \sum_{d=1}^{D}x_{u,d}x_{(|U|+i),d}\right)^2, \tag{5}$$

whose $G_E(\mathbf{X})$ is the Gauss-Newton matrix. To further simplify the formula for easier analysis, we let function Y denote the part of the inner part of (5). Then, the Y is as follows:

$$y_{u,i} = Y_{(u,i)}(\mathbf{X}) = \sum_{d=1}^{D}x_{u,d}x_{(|U|+i),d}, \tag{6}$$

where $y_{u,i}$, a scalar, denotes the inner product between $\mathbf{X_P}$ and $\mathbf{X_Q}$ Corresponding to $s_{u,i}$. Then, we have

$$E(\mathbf{Y}) = \frac{1}{2}\sum_{s_{u,i}\in K}(s_{u,i} - y_{u,i})^2. \tag{7}$$

where Y is a vector composed of K $y_{u,i}$'s. Next, we can compute $G_E(\mathbf{X})\mathbf{v}$, and $\omega_E$ can be derived as follows:

$$\omega_E = G_E(\mathbf{X})\mathbf{v} = J_Y^{\mathrm{T}}(\mathbf{X})\mathbf{I}J_Y(\mathbf{X})\mathbf{v}, \tag{8}$$

where $J_Y(\mathbf{X})$ represents the Jacobian matrix of **Y** w.r.t **X**, **I** is an identity matrix. By combining (4) and (8), we can get the $\omega_E$ as follows:

$$\begin{aligned}&\forall u \in U,\ d=1\sim D:\\ &\omega_{E_{u,d}} = \sum_{i\in K_u}\left(x_{(u+(|U|+i)),d}\left(\sum_{d=1}^{D}\left(v_{u,d}x_{(|U|+i),d} + x_{u,k}v_{(|U|+i),d}\right)\right)\right)\\ &\forall i \in I,\ d=1\sim D:\\ &\omega_{E_{(|U|+i),d}} = \sum_{u\in K_i}\left(x_{u,d}\left(\sum_{d=1}^{D}\left(v_{u,d}x_{(|U|+i),d} + x_{u,d}v_{(|U|+i),d}\right)\right)\right),\end{aligned} \tag{9}$$

where $\omega_{Ea,b}$ denotes the (a+b)-th element of $\omega_E$, $K_c$ represents the c-th user id or item id.

Moreover, With the incorporation of Tikhonov regularization term [23, 24], the objective function is as follows:

$$L(\mathbf{X}) = \frac{1}{2}\sum_{s_{u,i}\in K}\left(\left(s_{u,i} - \sum_{d=1}^{D}x_{u,d}x_{(|U|+i),d}\right)^2 + \lambda\sum_{d=1}^{D}\left(x_{u,d}^2 + x_{(|U|+i),d}^2\right)\right) \tag{10}$$
$$s.t.\ \forall u \in U, i \in I, d \in \{1,...,D\},$$

where $\lambda$, a hyperparameter, controls the effect of Tikhonov regularization term. We can obtain an increment $\Delta\mathbf{X}$ after solving (10), which is the update direction for **X**. Then, we can iteratively get the $\Delta\mathbf{X}$ for each iteration, the t-th update rule is as follows:

$$\mathbf{X}^{t+1} = \mathbf{X}^t + \Delta \mathbf{X}^t. \tag{11}$$

*B. A Distributed Particle Swarm Optimization*

Particle swarm optimization (PSO), a swarm intelligence algorithm, first proposed by Kennedy and Eberhart [30]. According to the basic definition of PSO algorithm, there are multiple populations and multiple particles, which can form different topological structures [31, 32]. Therefore, the PSO algorithm naturally has distributed parallelism. In order to speed up the execution of algorithm, as shown in Figure 1, we adopt a master-workers distributed topology. Each particle, a worker, represents an individual solution in D-dimension of the problem, and each particle has a different initial position, velocity, and its own historically optimal attractor. For example, particle $j$'s velocity $\mathbf{v}^t_j = (v^t_{j,1}, v^t_{j,2}, \ldots, v^t_{j,D})^T$ and position $\mathbf{p}^t_j = (p^t_{j,1}, p^t_{j,2}, \ldots, p^t_{j,D})^T$, where $v_{j,d} \in \left[v^{min}_{j,d}, v^{max}_{j,d}\right]$ and $p_{j,d} \in \left[p^{min}_{j,d}, p^{max}_{j,d}\right]$, and $1 \leq j \leq N$, $1 \leq d \leq D$, $v^{min}_{j,d}$ and $v^{max}_{j,d}$ are lower and upper bounds for velocity, $p^{min}_{j,d}$ and $p^{max}_{j,d}$ are lower and upper bounds for position, respectively. The master plays the role of managing the population and stores the global optimal attractor of the entire population.

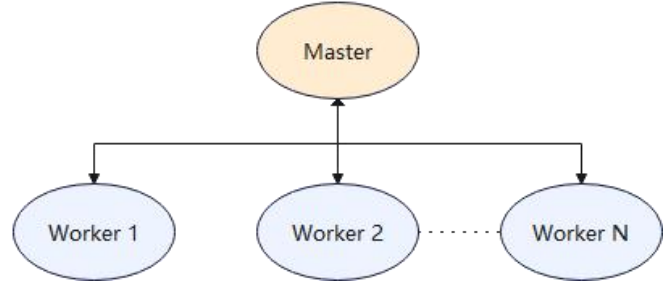

Fig. 1. Master-workers distributed topology.

After the initialization of the PSO, including each particle's speed and position. Each particle will search the space of the problem iteratively relying on their individual attractor and the global attractor. And the individual attractor and global attractor can make a trade-off in exploration and exploitation. In other words, they can balance the local search and global search capabilities [30-33]. In the $t$-th epoch, the update rule for each particle is as follows:

$$\begin{aligned} \mathbf{v}^t_j &= \omega \mathbf{v}^{t-1}_j + \Phi_1 \left(\mathbf{pb}^{t-1}_j - \mathbf{p}^{t-1}_j\right) + \Phi_2 \left(\mathbf{gb}^{t-1} - \mathbf{p}^{t-1}_j\right), \\ \mathbf{p}^{t+1}_j &= \mathbf{p}^t_j + \mathbf{v}^t_j, \end{aligned} \tag{12}$$

where $\omega$ refers to the inertia weigh, and its value range is [1,2], $\Phi_1 = c_1 r_1$ and $\Phi_2 = c_2 r_2$ are called individual learning factor and social learning factor, respectively. And individual attractor of each particle and the global attractor of the population are controlled by them separately. Among them, $c_1$, $c_2 > 0$, $r$ are independent random variables sampled from U(0, 1).

*C. An PSLF Model*

In general, hyperparameter optimization solves the following optimization problem [25]:

$$H^* = \arg\min F(L(\mathbf{X}), H), \tag{13}$$

where $H$ denotes the hyperparameters of $L(\mathbf{X})$, such as the regularization term $\lambda$ and the damping term $\gamma$, and $H^*$ denotes the best combination of hyperparameters.

Applying DPSO to solve (13), the position of $j$-th particle should be as follows:

$$p^t_j = \left(\lambda^t_j, \gamma^t_j\right)^T. \tag{14}$$

According to [23, 24] we set the value range of $\lambda$ and $\gamma$ to [0, 0.1] and [0, 300], respectively. The maximal velocity $v_{max}$ is set to 20% of the search range and $v_{min} = -v_{max}$.

In this paper, to evaluate the performance of PSLF model, we select root mean square error (RMSE) as the fitness value [33]. The evaluation metric is as follows:

$$Fitness = \sqrt{\left(\sum_{s_{u,i} \in \Gamma}\left(s_{u,i} - \sum_{d=1}^{D} x_{u,d} x_{(u+i),d}\right)^2\right) \Big/ |\Gamma|} \tag{15}$$

where Γ refers to the test set, |·| denotes the size of Γ, $s_{u,i}$ refers to the rating score of Γ. And fitness value will be calculated on the test set only.

Since each particle represents an SLF model with different hyperparameters. Not only can the combination of hyperparameters affect the performance of the SLF model, but the initial matrices of SLF model can also. We control each particle with the same initial matrices, thus ensure that the model performance is only affected by hyperparameters and nothing else, e.g. random seeds.

IV. EXPERIMENT

**Evaluation Metric**. In this paper, we choose RMSE as evaluation metric [12, 23, 24, 33]:

$$RMSE = \sqrt{\left(\sum_{s_{u,i} \in \Lambda}\left(s_{u,i} - \sum_{d=1}^{D} x_{u,d} x_{(|U|+i),d}\right)^2\right) \Big/ |\Lambda|}$$

where Λ denotes the test set or the validation set.

**Datasets**: There are three datasets utilized in this experiment, which were widely used in the previous LF model.

**D1**: MovieLens 1M dataset, a basic benchmark dataset for RS, is collected by GroupLens research [27, 34]. It includes 1,000,209 rating scores ranging from 1 to 5, which are generated by 6,040 MovieLens users on 3,900 movies. The data density of D1 is 4.25% only.

**D2**: Jester 1.1M dataset. D2, collected by joke-recommender Jester, contains 1,186,324 rating score [28]. These data are produced by 24,983 users who rated more than 36 jokes. The score of D2 is at the range of [-10, 10] and its data density is 47.32%.

**D3**: WS-Dream response time dataset. Real-world QoS evaluation results from 339 users on 5,825 Web services [36-40]. D2 is widely used for QoS prediction, and is widely used to test the performance of LF models. It holds 1,873,828 response data as rating scores.

**Implementation Details.** For all experiments, we adopt five-fold cross-validation. All datasets are divided into three parts, i.e., training set, test set and validation set, with a ratio of 6:2:2. The training set will be used to train SLF model, and the test set will be used by DPSO to calculate the fitness value to guide the evolution of hyperparameters. The validation set will be used to verify the generalization error of the fine-tune PSLF model. All experiments are implemented by using a machine with an Intel-Xeon 6230 CPU, Nvidia Tesla T4 and 512GB DDR4 RAM. All details about the models are implemented with OpenJDK 11.

**As shown in Table I-III, compared with M2-M4, on the same dataset, M1 is much better in representational power and computational time cost.** From Table II, we can compare the performance difference of M1-M4 by calculating the RMSE, lower RMSE means higher prediction accuracy. For example, on D1, the RMSE of M1 is 1.17%, 1.87%, 3.42% lower than that of M2-M4, respectively. It means that the representation ability of M1 in these models, M1-M4, is the optimal.

Table I. Details of Compared Models

| No. | Name | Description |
|---|---|---|
| M1 | PSLF | The model proposed in this paper |
| M2 | NLF | An NLF model proposed in [8] |
| M3 | ANLF | An ANLF model proposed in [31] |
| M4 | I-AutoRec | An I-AutoRec model proposed in [42] |

TABLE II. PERFORMANCE OF DIFFERENT MODELS IN RMSE

| Model | D1 | D2 | D3 |
|---|---|---|---|
| M1 | **0.85367±0.00015** | **1.01284±0.00114** | **1.10738±0.00009** |
| M2 | 0.86386±0.00030 | 1.02305±0.00186 | 1.11791±0.00369 |
| M3 | 0.86993±0.00067 | 1.02971±0.00151 | 1.13891±0.00222 |
| M4 | 0.88392±0.00113 | 1.15428±0.00163 | 1.16793±0.00077 |

TABLE III. TOTAL TIME OF DIFFERENT MODELS(SECOND)

| Model | D1 | D2 | D3 |
|---|---|---|---|
| M1 | **376±145** | **634±223** | **166±37** |
| M2 | 3073±542 | 1752±147 | 1303±37 |
| M3 | 3742±1999 | 3309±1645 | 3738±1483 |
| M4 | 7143±22 | 1308±16 | 5220±27 |

## V. CONCLUSION

In this paper, we propose a hyperparameter-adaptive second-order LF model called PSLF. To achieve this goal, we incorporate DPSO into the SLF model's hyperparameters optimization stage. The PSLF model, free from the trouble of hyperparameters tuning, has significant advantages in computation time compared with the SLF model tuned by grid search or trial-and-error method, and has similar convergence accuracy. However, in the PSLF model, vanilla PSO algorithm is taken, so this may lead to premature convergence. Whether other PSO variants can avoid this problem and get a better combination of hyperparameters Meanwhile, hessian-free optimization will cause huge computational overhead, which is very obvious on large-scale dataset. Whether other better second-order approximate optimization algorithms for large-scale datasets in the LF model. We plan to solve these issues in the future.